\documentclass[11pt,a4paper]{article}
\usepackage[hyperref]{acl2020}
\usepackage{times}
\usepackage{latexsym}
\usepackage{todonotes}
\usepackage{hyperref}

\newcommand{\suzan}[1]{\textcolor{black}{#1}}
\newcommand{\sv}[1]{\textcolor{black}{#1}}

\usepackage{microtype}

\aclfinalcopy 

\title{Small data problems in political research: a critical replication study }

\author{Hugo de Vos \\
  Institute of Public Administration \\
  Leiden University \\
  \texttt{h.p.de.vos@fgga.leidenuniv.nl} \\\And
  Suzan Verberne \\
  Leiden Institute of Advanced \\ Computer Science (LIACS) \\
  Leiden University \\
  \texttt{s.verberne@liacs.leidenuniv.nl} \\}

\date{\today}

\begin{document}
\maketitle
\begin{abstract}
In an often-cited 2019 paper on the use of machine learning in political research, Anastasopoulos \& Whitford (A\&W) propose a text classification method for tweets related to organizational reputation. The aim of their paper was to provide a `guide to practice' for public administration scholars and practitioners on the use of machine learning. 
In the current paper we follow up on that work with a replication of A\&W's experiments and additional analyses on model stability and the effects of preprocessing, both in relation to the small data size. We show that (1) the small data causes the classification model to be highly sensitive to variations in the random train--test split (2) the applied preprocessing causes the data to be extremely sparse, with the majority of items in the data having at most two non-zero lexical features. \suzan{With additional experiments in which we vary the steps of the preprocessing pipeline, we show that the small data size keeps causing problems, irrespective of the preprocessing choices.} 
Based on our findings, we argue that A\&W's conclusions regarding the automated classification of organizational reputation tweets -- either substantive or methodological – can not be maintained and require a larger data set for training and more careful validation.

\end{abstract}

\section{Introduction}\label{sec:intro}

In\footnote{All data and scripts are published at: \url{ https://anonymous.4open.science/r/Critical_Replication_ML_in_PA-3F20/README.md }} 2019, \suzan{the Journal of Public Administration Research and Theory (JPART) published a paper on the use of Machine Learning (ML) in political research}~\cite{anastasopoulos_machine_2019} (A\&W). With this paper, A\&W attempt `to fill this gap in the literature through providing an ML ``guide to practice'' for public administration scholars and practitioners'~\cite[p. 491]{anastasopoulos_machine_2019}.
A\&W present an example study, in which they aim to `demonstrate how ML techniques can help us learn about organizational reputation in federal agencies through an illustrated example using tweets from 13 executive federal agencies'~\cite[p. 491]{anastasopoulos_machine_2019}. In the study, a model was trained to automatically classify whether a tweet is about moral reputation or not. According to the definition scheme by A\&W, a tweet addresses moral reputation if it expresses whether the agency that is tweeting is compassionate, flexible, and honest, or whether the agency protects the interests of its clients, constituencies, and members~\cite[p. 509]{anastasopoulos_machine_2019}. The conclusion of the example study was that `the Department of Veterans Affairs and the Department of Education stand out as containing the highest percentage of tweets expressing moral reputation.'~\cite[p. 505]{anastasopoulos_machine_2019}. 

A\&W also provided a concise, but more general, introduction to machine learning for Public Administration scientists, of which the example study was an integral part illustrating how machine learning studies could work.
The concise overview on supervised machine learning makes the paper a valuable addition to the expanding literature on machine learning methods in political research. However, the example study contains several shortcomings that are not addressed by A\&W. \sv{A possible undesired result is} that practitioners or researchers unfamiliar with machine learning will follow the wrong example and consequently conduct a flawed study themselves. It is for this reason that we zoom in on the data used in the example study and the validation that is reported by A\&W\sv{, showing the problems with their study}.

A\&W train a Gradient Boosted Tree model with bag-of-words features on the binary classification task to recognize whether a tweet is about moral reputation or not. The model is first trained on a data set of 200 human-labeled tweets and evaluated using a random 70-30 train--test split. The trained model is then used to automatically infer a label for 26,402 tweets. Based on this larger data set, A\&W analyze to what extent specific US institutions work on their moral reputation via Twitter.  

The core problem with this set-up is that the training data set is too small to train a good model. We show that this results in a model that is of drastically different \suzan{quality when the random split of the data is varied}, an effect that we will call model (in)stability.
The consequences of these mistakes are that the model by A\&W can not reliably be used for data labeling, because data generated with this model can not be assumed to be correct. 
Although the mistakes can only be solved with a larger data set, the flaws could have been detected if the model would have been validated more thoroughly by the authors. 

The consequences for the conclusions in the \sv{A\&W paper itself} might be relatively small, because it is only one example without overly strong substantive claims. However, more importantly, the weaknesses of the paper might also influence any future research based on the study; the paper was published in a high-impact journal and has been cited 49 times since 2019.\footnote{According to Google Scholar, June 2021}

\sv{In this paper, we} replicate the results by A\&W, and analyze their validity. We perform what \citet{belz2021systematic} call a \textit{reproduction under varied conditions}: a reproduction where we ``deliberately vary one or more aspects of system, data or evaluation in order to explore if similar results can be obtained" (p. 4). 
We show that the A\&W results can indeed be reproduced, yet only in very specific circumstances (with specific random seeds). We demonstrate that the methods have flaws related to data size and quality, which lead to model instability and data sparseness. This means that the `guide to practice' \suzan{that A\&W aim to provide requires careful attention by any follow-up work.}

We address the following research questions:

\begin{enumerate}
\item \suzan{What is the effect of small training data on the stability of a model for tweet classification?} 
\item \suzan{To what extent do changes in the preprocessing pipeline influence the model quality and stability in combination with the small data size}?
\end{enumerate}

We first make a comparison between the data set of A\&W and other text classification studies in the political domain (Section~\ref{sec:related_work}). We then report on the replication of A\&W's results, followed by an analysis of the model stability under the influence of different random data splits (Section~\ref{sec:modelstability}). \suzan{In Section~\ref{sec:smalldata} we conduct additional experiments varying the preprocessing pipeline to further analyze} the implications of the small data size on the usefulness of the data for the classification task. We conclude with our recommendations in Section~\ref{sec:conclusions}.  

\section{Related work on political text classification and data size}\label{sec:related_work}

In the field of political science, text mining methods (or Quantitative Text Analysis (QTA) as it is called in the Political Science community) have been used for about a decade. One of the first major papers on the use of automatic text analysis in the field was \citet{grimmer_text_2013}. In this seminal paper the pros and cons of using automatic text analysis are discussed.

Another major contribution to the field is the Quanteda package~\cite{benoit2018quanteda} in R. This R package contains many tools for Quantitative Text Analysis such as tokenization, stemming and stop word removal and works well with other (machine learning) R packages like topicmodels~\cite{grun2021topicmodels} and xgboost~\cite{chen_xgboost_2016}. This package that has been developed by and for Political Scientists and Economists has already been widely used in the community. 

A\&W used the tm package~\cite{feinerer2020tm} for text mining in R. The data set used to train their machine learning model consists of a total of two hundred tweets. Eighty two of those were manually labeled by the authors as being about moral reputation and 118 as not being about moral reputation.\footnote{Originally, they also had the tweets annotated via crowd sourcing, but the resulting annotations had such a low inter-coder reliability that they decide not to used them due to the poor quality.} The average length of a tweet in the data set is $17.7$ words with a standard deviation of $4.4$.

In comparison to other studies that used machine learning for tweet classification, 200 tweets is notably small. 
\sv{The issue of the small data size is aggravated by the short length of tweets: }
They contain few words compared to other document types such as party manifestos~\cite{merz_manifesto_2016,verberne_automatic_2014} or internet articles~\cite{fraussen_assessing_2018}. Because tweets are so short, the bag-of-words representation will be sparse, and in a small data set many terms will only occur in one or two tweets. This makes it difficult to train a generalizable model, as we will demonstrate in Section~\ref{sec:smalldata}.

\suzan{Based on the literature, there is no} clear-cut answer to how much training data is needed in a text classification task. This depends on many variables, including the text length, the number of classes and the complexity of the task. Therefore we can not say how many tweets would have sufficed for the goal of A\&W. What is clear from \suzan{related work}, is that it should be at least an order of magnitude larger than 200. 
\citet{elghazaly_political_2016}, for example, used a set of 18,278 hand-labeled tweets to train a model for recognizing political sentiment on Twitter. 
\citet{burnap_cyber_2015} used a set of 2,000 labeled tweets to train a model that classifies the offensiveness of Twitter messages.
\citet{amador_diaz_lopez_predicting_2017} used a total of 116,866 labeled tweets to classify a tweet about Brexit as being Remain/Not Remain or Leave/Not Leave. 

Most, if not all, of the recent work in the field of computational linguistics uses transfer learning from large pre-trained language models for tweet classification, in particular BERT-based models~\cite{devlin2018bert}. In these architectures, tweets can be represented as denser vectors, and the linguistic knowledge from the pretrained language model is used \sv{for representation learning}. The pretrained model is finetuned on a task-specific dataset, which in most studies is still quite large. \citet{nikolov2019nikolov}, for example, used a training set of 13,240 tweets \cite{zampieri2019predicting} to \sv{fine-tune a BERT} model to classify the offensiveness of a tweet. This resulted in an accuracy of 0.85.

A more general point of reference about sample sizes for tweet classification is the SemEval shared task, a yearly recurring competition for text classification often containing a Twitter classification task. 
For example, in 2017 there was a binary sentiment analysis task where participants could use a data set of at least\footnote{There were other tasks where more training data was available.} 20,000 tweets to train a model~\cite{rosenthal_semeval-2017_2019}. 

These studies show that even in binary classification tasks using twitter data, a lot of data is often needed to achieve good results, despite that those tasks might look simple at first glance. \suzan{In the next section, we empirically show that the A\&W data is too small for reliable classification.}

\section{Replication and model stability} \label{sec:modelstability}

A\&W report good results for the classifier effectiveness: a precision of 86.7\% for the positive class (`about moral reputation'). 
In this section we present the results of an experiment that we did to validate the reported results. In addition to that we will also assess the stability of the model. By this we mean how much the model and its performance changes when the data is split differently into a train and  test set. We argue that if an arbitrary change (like train test split) leads to big changes in the model, the generalizability of the model is poor, \suzan{because it shows that changes in data sampling results in changes in model quality, and hence in different classification output}.

\subsection{Exact replication} We first completed an exact replication of the experiment of A\&W to make sure we started from the same point. We followed the data analysis steps described in A\&W exactly. 
Thanks to the availability of the data and code, the study could be replicated with ease.
The exact replication yielded the same results as reported in A\&W.  


\subsection{Varying the random seed} 

In their experiments A\&W 
\sv{make a random 70-30 train--test split of the 200 labelled tweets:}
140 tweets are randomly sampled 
to be the train set 
and the remaining 60 tweets form the test set. In \sv{their paper, they present} the result of only a single random split. For reproducibility reasons A\&W use a single random seed for the train--test split.\footnote{In their case this seed is $41616$}

In order to assess the generalizability of the model, we generated a series of one thousand random seeds (the numbers 1 to 1000). This resulted in a thousand different train-/test splits of the tweets. We reran the experiment by A\&W with all the random train--test splits, keeping all other settings unchanged. In all cases, the train set contained 70\% (140) of the labeled tweets and the test set 30\% (60) of the labeled tweets. For each of the thousand runs we calculated the precision, in the same way that A\&W did. 

If a model is robust, most of the different configurations should yield approximately the same precision. Inevitably, there will be some spread in the performance of the models but they should group closely around the mean precision which indicates the expected \sv{precision} on unseen data. 

\subsection{Results of varying the random seed}

\begin{figure*}[t]
\includegraphics[width=\linewidth]{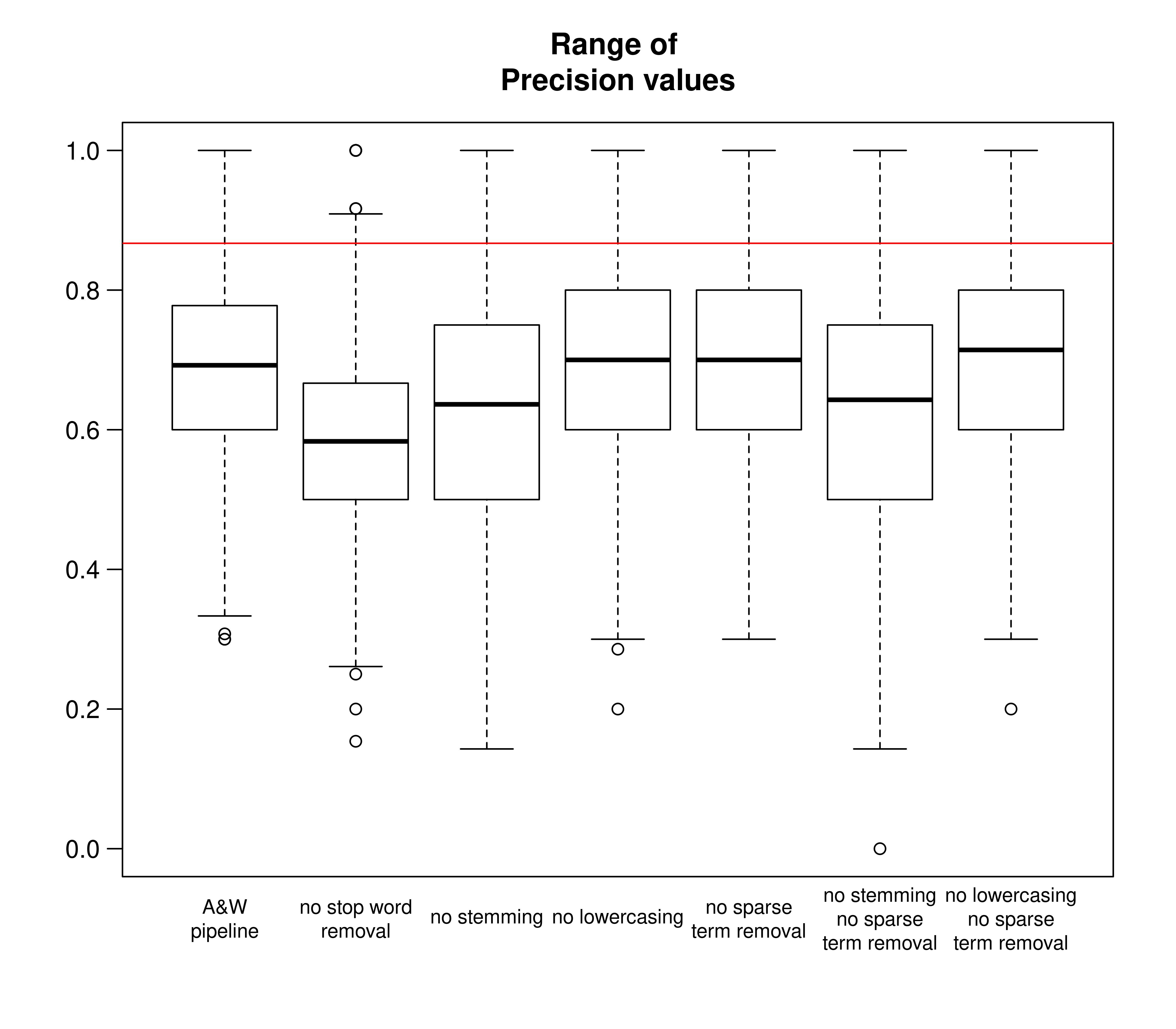}
\caption{A visualization of the spread of results of the random seed variation experiment. The leftmost box summarizes the results of 1000 different runs with the same settings as A\&W, except for the random seeds. The \suzan{horizontal red} line depicts the precision that is reported by A\&W. The other box plots are the results of 1000 runs where each time one preprocessing step is omitted as described in section \ref{subsec:preprocessing_differences}.} \label{fig:boxplot}
\end{figure*}

Our experiment resulted in precision scores that ranged from 0.3 to 1.0. The mean precision was $0.67$ with a standard deviation of $0.14$. The median was $0.69$. The mean and standard deviations of the 1000 runs \sv{for precision, recall and F1} are listed in Table \ref{tab:replication_stats}. The distribution of precision values is also depicted in the leftmost boxplot in Figure \ref{fig:boxplot}. \sv{The table indicates that the model on average performs rather poorly for a binary classification task: the F-score for the positive class is 0.40 and for the negative class 0.75. In addition,} the plot as well as the standard deviations in the table show \sv{a large variance in quality between different random seeds}. This indicates that the model is 
unstable.

\begin{table}[h]
\begin{tabular}{l|rl|}
\cline{2-3}
                                     & \multicolumn{1}{l}{Class}        &             \\
                                     & \multicolumn{1}{l}{Positive}     & Negative    \\ \hline
\multicolumn{1}{|l|}{Precision (sd)} & \multicolumn{1}{r|}{0.69 (0.14)} & 0.65 (0.06) \\
\multicolumn{1}{|l|}{Recall (sd)}    & \multicolumn{1}{r|}{0.30 (0.10)} & 0.90 (0.08) \\
\multicolumn{1}{|l|}{F1-score (sd)}  & \multicolumn{1}{r|}{0.40 (0.09)} & 0.75 (0.05) \\ \hline
\end{tabular}
\caption{\label{tab:replication_stats} The means and standard deviation for the evaluation statistics.}
\end{table}

What also stands out is that the result by A\&W (the horizontal red line in Figure~\ref{fig:boxplot}) appears to be exceptionally high. Out of the 1000 runs, only 6 were able to match or outperform the precision presented in A\&W (.867). The mean precision over 1000 runs is much lower than the precision reported by A\&W. We argue that the mean precision over 1000 runs is more likely to be \suzan{a realistic reflection of the actual model precision than the result for one random split.}

From these results, we conclude that the model \sv{quality is relatively poor and} unstable: changing the train--test split, an arbitrary alteration that should not make a big difference, leads to a wide range of outcomes. This has an effect on the generalizing power of the machine learning model: Although the reported results on the test set (with only one particular random seed) are good, they are not generalizable to other data splits. 

That the model generalizes poorly is in fact confirmed by Figures 3 and 5 in~\citet[p. 503 and 506]{anastasopoulos_machine_2019}. These figures show that solely the occurrence of the word `learn' or `veteran' will make the model predict that a tweet is about moral reputation, regardless of any other words occurring in the tweet. This is an effect of these words being overrepresented in the data sample. This artefact effect is more likely to occur if a data sample is too small. This situation will lead to overfitting of the model\sv{, a likely effect that is not described by A\&W}. We explore the effects of the small data size in more detail in the next section.

\section{Implications of small data sets on data quality }\label{sec:smalldata}
In the previous section we showed how the small amount of data leads to poor model stability. 
In this section we show how the small number of tweets negatively affects the quality of the data set that serves as input to the machine learning model. \suzan{We also experiment with other preprocessing choices to investigate the effect on the model quality and stability.}

A\&W apply a number of common preprocessing steps to their data: \begin{itemize} 
\item Decapitalisation (e.g. `Veteran' $\rightarrow$ `veteran') 
\item Removal of all special characters, numbers,  punctuation, and URLs
\item Stop-word removal
\item Removal of rare terms: all words that occur in fewer than 2\% of the tweets are removed from the data.
\item Stemming with the SnowballC stemmer \cite{bouchetvalat2020Snowball}
\end{itemize}
The remaining unigrams are used as count features in the bag-of-words model.

In the next two subsections, we first analyze the effect of word removal (stop word and rare words), and then investigate the effect of changing the preprocessing steps on the quality of the model.

\subsection{The effect of removing words}
As introduced above, A\&W remove both stop words and rare words from the data before the document--term matrix is created. Examples of stop-words removed by A\&W are `they', `are', `is' and `and'. Removing such words prevents a model from learning that, for example, the word `the' signals that a tweet is about moral reputation because the word `the' occurs, by chance, more often in tweets about moral reputation. 

Similarly, rare words are not considered to be a relevant signal. For example, the word `memorabilia' occurs only one time in the tweet collection of A\&W, and this happens to be in a tweet about moral reputation. 
A machine learning algorithm could, therefore, infer that `memorabilia' contributes positively to a tweet being about moral reputation, which is not a generalizable rule. For this reason words that occur only rarely are commonly removed, as do A\&W.

However in combination with the small data size, the effect is that almost every word is either a stop-word or a rare word. Consequently, removing stop words and rare words leads to tweets from which almost every word is deleted. 
In fact, in the preprocessing setting of A\&W, 95\% of all the tokens in the collection were removed, reducing the dictionary size from 1473 to 70. 
As a result, many tweets have fewer than three \suzan{non-zero features}, making it \suzan{difficult for the model} to predict the label of those tweets. 

This effect is further illustrated in Table~\ref{tab:wordsintweets}, which lists the number of tweets from the data set with a given number of words. This table shows that after removing rare words and stop words, 15\% of the tweets in the collection have \suzan{no non-zero features at all}, and 24\% percent \suzan{are represented by} only one \suzan{non-zero feature}. As a result of this, the model tried to learn how to recognize whether a tweet is about moral reputation or not based on tweets with barely any words in them.

The situation is even more clear in the unlabeled collection. In this set, from 25\% of the tweets every word was removed. By coincidence, the model in A\&W learned that every tweet with no words left was about moral reputation. This means that 25\% of the data set on which A\&W based their conclusion, has received the label `about moral reputation', while this is impossible to say based on zero words. This means that at least 25\% of the tweets' labels can not be trusted.

\begin{table*}[t]
\begin{tabular}{|p{2cm}|p{1cm}|p{1cm}|p{1cm}|p{1cm}|p{1cm}|p{1cm}|p{1cm}|p{1cm}|p{1cm}|}
\hline
N  & 0  & 1   & 2   & 3   & 4   & 5   & 6   & 7   & 8 \\
\hline
Coded set  & 25 (15\%)  & 47 (24\%)  & 52 (26\%)   & 37 (19\%)   & 13 (7\%)   & 11 (6\%)  & 4 (2\%)   & 4 (2\%)  & 1 (0.05\%)  \\
\hline
Uncoded set  & 6519 (25\%)  & 8099 (31\%)  & 6295 (21\%)  & 3558 (13\%)  & 1349 (5\%)  & 441 (1.7\%)  & 108 (0.4\%)  & 30 (0.1\%)  & -- \\
\hline
\end{tabular}
\caption{\label{tab:wordsintweets} The amount and proportion of tweets from the human-labeled set and the uncoded set that contain N words.}
\end{table*}

The instability can be clarified further with a few examples. Example 1 (a tweet by @USTreasury with the label `not about moral reputation') has only the words `new' and `provides' left after preprocessing. 
From example 2 (by @USDOT with the label `not about moral reputation') only the word `today' is left. Example 3 (by @CommerceGov) is `about moral reputation' and only the word `learn' is left.

\begin{enumerate}
    \item \textbf{Before preprocessing}: ``We have a new mobile website that provides a virtual tour of 1500 Penn \textless url\textgreater  \textless url\textgreater `' \\
    \textbf{After preprocesing}: ``new provides''
    \item \textbf{Before preprocessing}: ``RT @SenateCommerce TODAY AT 10AM @SenateCommerce to hold a hearing to examine \#InfrastructureInAmerica with testimony from @SecElaineChao''\\
    \textbf{After preprocessing}: ``today''
    \item \textbf{Before preprocessing}: ``RT @NASA: We've partnered with @American\_Girl to share the excitement of space and inspire young girls to learn about science, technology,..."\\
    \textbf{After preprocessing}: ``learn''
\end{enumerate}

It is difficult -- if not impossible -- to train a reliable model on these very limited representations of tweets. 

This could have been prevented if the number of tweets would have been larger. As a consequence of Heaps' law, the number of new unique terms becomes smaller with every new document that is added ~\cite{heaps1978information}. As a result of this, a document collection with more documents/tweets will have fewer rare terms.

\subsection{The effect of preprocessing differences}\label{subsec:preprocessing_differences}

We investigated what the effect on the quality of the model is of different preprocessing choices. We created variants of A\&W's pipeline with one of the following adaptations:
\begin{itemize}
    \item Not removing stopwords
    \item No stemming
    \item No lowercasing
    \item Not removing rare words
    \item No stemming and not removing rare words
    \item No lowercasing and not removing rare words
\end{itemize}

Like in Section~\ref{sec:modelstability} we ran each model 1000 times with different random seeds and show the range of precision values for each setting in Figure~\ref{fig:boxplot}. This shows that there \sv{are} differences \sv{between the} preprocessing settings\suzan{, but the model remains highly unstable and has relatively low median precision scores between 0.59 and 0.71 for the different preprocessing choices.}

\begin{table*}[t]
\begin{tabular}{l|cc|cc|}
\cline{2-5}
                                                           & \multicolumn{2}{c|}{Dict size}                                                                                                                               & \multicolumn{2}{c|}{\begin{tabular}[c]{@{}c@{}}\% of tweets with n terms\\ after rare term removal\end{tabular}} \\ \hline
\multicolumn{1}{|l|}{experiment}                           & \multicolumn{1}{c|}{\begin{tabular}[c]{@{}c@{}}before rare \\ term removal\end{tabular}} & \begin{tabular}[c]{@{}c@{}}after rare\\ term removal\end{tabular} & \multicolumn{1}{c|}{0 terms}                                       & 1 term                                      \\ \hline
\multicolumn{1}{|l|}{A\&W}                                 & 1473                                                                                     & 70                                                                & 15 \%                                                                 & 24 \%                                          \\
\hline
\multicolumn{1}{|l|}{No stopword removal}                  & 1529                                                                                     & 96                                                                & 2 \%                                                                  & 8 \%                                           \\
\multicolumn{1}{|l|}{No stemming}                          & 1623                                                                                     & 47                                                                & 25 \%                                                                 & 35 \%                                          \\
\multicolumn{1}{|l|}{No lowercasing}                       & 1515                                                                                     & 73                                                                & 13 \%                                                                 & 25 \%                                          \\
\multicolumn{1}{|l|}{No rare term removal}                 & 1473                                                                                     & NA                                                              & NA                                                                   & NA                                            \\
\multicolumn{1}{|l|}{No stemming and rare term removal}    & 1623                                                                                     & NA                                                              & NA                                                                   & NA                                            \\
\multicolumn{1}{|l|}{No lowercasing and rare term removal} & 1515                                                                                     & NA                                                              & NA                                                                   & NA                                            \\ \hline
\end{tabular}
\caption{\label{tab:dictsizes} The size of the dictionary as the result of omitting different preprocessing steps before and after the removal of rare terms. Also the percentage of tweets with 0 and 1 terms after rare term removal is listed. }
\end{table*}

The different preprocessing steps naturally lead to different dictionary sizes (The number of variables in the document--term matrix). 
Not lowercasing, for example, increases the number of terms in the dictionary, as words like `veteran' and `Veteran' are now seen as diferent tokens. The effect of the different preprocessing steps on the dictionary sizes is listed in Table~\ref{tab:dictsizes}.

Table~\ref{tab:dictsizes} shows that omitting any of the preprocessing steps (except rare term removal) increases the dictionary size. This makes sense, because all those steps are designed to reduce the dictionary size by collating \suzan{different word forms to one feature or removing words.} In the case of no stopword removal, the dictionary size after rare term removal is larger than if the pipeline of A\&W is applied. This can be explained since the stopwords that remain, are never rare terms and thus are not removed. This also explains why there are almost no tweets with only 0 or 1 terms in this setting, because almost every tweet contains a stopword.

Omitting the stemming procedure leads to a larger dictionary size before, but a smaller dictionary size after rare term removal. Because terms are not collated, there will be more unique terms, but all those terms are more likely to be rare. The effect of more terms being removed also shows in the large amount of tweets with 0 or 1 term. \suzan{The effect that 60\% of the tweets only contains 0 or 1 words (25+35\%) explains why the settings without stemming are the least stable settings of all (Figure~\ref{fig:boxplot}).}

Not lowercasing the tweets only seems to have a marginal effect. This is likely due to the fact that the number of (non rare) words starting with a capital letter is already small to begin with.

\sv{In conclusion,} Figure~\ref{fig:boxplot} shows that \suzan{the effect of preprocessing choices has on the precision is relatively small,} if anything omitting the preprocessing steps made the models worse on average. 
\suzan{This confirms that the data set size is detrimental to the model quality -- even after lowercasing, stemming, removing stopwords and rare words, the model can not generalize between different data sampling splits. }

\section{Conclusions} \label{sec:conclusions}

In this paper, we replicated and analyzed a study that was published in JPART that explains and illustrates how to \sv{use machine learning for analyzing Twitter data.} 
The data set used in the example study was too small to train a reliable model. We demonstrated this with a number of experiments: First, we replicated the example study exactly, then we studied the stability of the model by varying the train--test split. In the final experiment, we \suzan{analyzed the effect of different preprocessing choices on } 
the quality of the data and, subsequently, the quality of the model.

\paragraph{Answers to research questions} We found that the results by A\&W could be replicated, but only under very specific conditions; \suzan{our experiment with 1000 random train--test splits showed that only 6 of those 1000 splits could meet or outperform the precision reported by A\&W. We find a median precision of 69\%, as opposed to the 86.7\% reported by A\&W. }
\suzan{In response to RQ1, what the effect of small training data on the stability of a model for tweet classification is, we show that the small data size has caused the model to be highly unstable, with precision scores ranging from 30\% to 100\% depending on the train--test split used.}

\suzan{We analyzed the effect of choices in the preprocessing pipeline by varying them. In each setting, the range of precision scores obtained in 1000 train--test splits was large and none of the settings could improve upon the A\&W setting. In response to RQ2, to what extent changes in the preprocessing pipeline influence the model quality and stability, we show that the effect of preprocessing choices is relatively small; we obtain median precision scores between 59\% and 71\% with large standard deviations. We conclude that the data set is too small to train a stable, high-quality model, largely irrespective of the preprocessing steps. }

Overall, we showed that the small data issues reduce the validity of the results reported in A\&W, especially as a machine learning example for the \sv{political research} community.

\paragraph{Recommendations for future work} \suzan{As discussed in Section~\ref{sec:related_work}}, there is no golden rule for how much training data is needed. In general; the shorter a document is, the more documents you need in the training set. In the case of tweets, one would need at least a few thousand hand-labeled training examples. Also, it is important to always report the size of the data set. Not only the number of documents/tweets but also the average number of words in each document.

Apart from recommendations on data set size, we also showed that validation of the model stability can be done by varying the random seed. This can indicate whether more training data is needed for a reliable classifier. 

Any researchers seeking to follow up on A\&W in designing a machine learning study could additionally consult \citet{lones2021avoid}, a concise overview of a multitude of points to consider to avoid machine learning pitfalls.\\

Finally, we would like to stress the importance of replication and reproducability. As is noted in \citet{cohen2018three} and \citet{belz2021systematic} replication studies in NLP are becoming more common in recent years. \citet{belz2021systematic} conclude that ``worryingly small differences in code have been found to result in big differences in performance." (p. 5). This statement is reinforced by the findings in our paper.

A precondition for good \sv{debates in social and political sciences} based on the outcomes of NLP experiments is that those outcomes are demonstrably reliable. If the results are not robust, a further debate \sv{based on the implications of the results} is pointless.

\bibliography{acl2020}
\bibliographystyle{acl_natbib}

\end{document}